\documentclass[fleqn,10pt]{SelfArx} 

\usepackage{lipsum} 
\usepackage[numbers]{natbib}
\usepackage{subcaption}
\usepackage{float}

\definecolor{color1}{RGB}{0,0,90} 
\definecolor{color2}{RGB}{0,20,20} 

\usepackage{hyperref} 
\hypersetup{hidelinks,colorlinks,breaklinks=true,urlcolor=color2,citecolor=color1,linkcolor=color1,bookmarksopen=false,pdftitle={Title},pdfauthor={Author}}

\PaperTitle{Short Run Transit Route Planning Decision Support System Using a Deep Learning-Based Weighted Graph} 

\Authors{Nadav Shalit\textsuperscript{1}*, Michael Fire\textsuperscript{1}, Dima Kagan\textsuperscript{1}, Eran Ben-Elia\textsuperscript{1}} 
\affiliation{\textsuperscript{1}\textit{Ben-Gurion University}} 
\affiliation{*\textbf{Corresponding author}: shanad@post.bgu.ac.il} 

\newcommand{\keywordname}{Keywords} 

\Abstract{
Public transport routing plays a crucial role in transit network design, ensuring a satisfactory level of service for passengers. 
However, current routing solutions rely on traditional operational research heuristics, which can be time-consuming to implement and lack the ability to provide quick solutions.
Here, we propose a novel deep learning-based methodology for a decision support system that enables public transport (PT) planners to identify short-term route improvements rapidly.
By seamlessly adjusting specific sections of routes between two stops during specific times of the day, our method effectively reduces times and enhances PT services.
Leveraging diverse data sources such as GTFS and smart card data, we extract features and model the transportation network as a directed graph. 
Using self-supervision, we train a deep learning model for predicting lateness values for road segments.
 These lateness values are then utilized as edge weights in the transportation graph, enabling efficient path searching.
Through evaluating the method on Tel Aviv, we are able to reduce times on more than 9\% of the routes. 
The improved routes included both intraurban and suburban routes showcasing a fact highlighting the model's versatility. 
The findings emphasize the potential of our data-driven decision support system to enhance public transport and city logistics, promoting greater efficiency and reliability in PT services.}

\begin{document}

\maketitle 



\section*{Introduction}
Migration and natural growth during the last century have contributed to rapid urbanization worldwide as people seek better employment and living conditions. The growth of the urban population has resulted in the exasperation of traffic problems and growth in the number of cars on urban roads, raising the need for more sustainable forms of mobility. This problem has led to an increase in the significance and importance of transport planning \citep{emberger2008ideal},  including decisions on investment in improving and expanding public transport (PT) services~\citep{petrovic2016appraisal, al2011composite}.

While PT is one solution to improve congestion and climate change, the quality of provided services must be improved to attract new passengers \citep{vitale2014decision} and to steer passengers toward energy and space-efficient transportation modes \citep{cascetta2014quality}. Therefore, for smart and sustainable cities, PT planners should identify best practices in route planning \citep{mcleod2017urban}. The PT network is a complex system comprised of stops, vehicles, routes, and other temporal and spatial elements \citep{ceder2016public}, operating scheduled vehicle trips available to all paying passengers and capable of servicing very different origins and destinations. However, to attract loyal ridership a PT system should also be punctual and regular \citep{walker2012human}. For example, in Israel (an official OECD member), buses accounted nationally for 85\% of PT trips in 2019, with more than 2M passengers served daily while the country suffers from a shortage of adequate PT infrastructure (namely too few priority lanes - 14 meters per capita, compared to 300 meters in the EU), thus resulting in poor PT service punctuality. Although understanding the patterns of PT use is crucial to its planning, this task remains a major challenge in practice and research \citep{li2018smart}.

Many algorithms and methods were suggested to deal with various aspects of PT planning.
One common way to study PT planning is to model the problem as a network.
\citet{wang2015efficient} proposed Timetable Labelling (TTL), a method for route planning using timetable graphs.
They save inside the graph data representing the shortest travel time for various vehicles at a specific departure time.
Using this data, they present an efficient method for querying transportation paths.
Another direction that researchers have taken to improve public transportation is traffic prediction.
\citet{barnes2020bustr} proposed BusTr, a method for estimating bus travel time between two stops with a predefined route using contextual features.
\citet{tang2022domain} developed a deep-learning framework for transferring forecasts across cities by fine-tuning the network.
This way, they allow cities with small amounts of data to produce accurate forecasts.

While many papers deal with predicting traffic, bus arrivals, and route planning, there are not enough studies that detect short-term improvements that decision-makers can immediately incorporate. 
In this paper, we propose a novel methodology for establishing a decision support system (DSS) using deep learning and graph theory for short-term PT route planning that optimizes routes for expected delays in their schedules. Our method utilizes smart card data that best represents the real-world state of PT. 
Our methods allow changing routes between two stops at a specific time of the day to reduce commute time.
Since the stops and their order stay the same and only the route changes, decision-makers can easily use our method to improve PT services.

Our method involves a process incorporating several key steps. First, data is collected from various sources, including General Transit Feed Specification (GTFS) and smart cards, encompassing information about passenger alighting, road networks, intersections, stops, and other relevant elements. Second, we construct a network representation wherein road segments serve as edges, while intersections and stops are represented as nodes.  The next step extracts a range of features containing geospatial information, temporal data, and other pertinent variables from the collected datasets. These features are then assigned as attributes to the corresponding edges in the network, enabling comprehensive characterization of the transportation system. Next, a deep learning model is trained using self-supervision, leveraging the extracted features. 
This model is fine-tuned to predict the lateness value associated with each road segment.
In our case, lateness is the temporal difference between the planned schedule GTFS and the actual boarding time from smart card data.   
Subsequently, the lateness values obtained from the trained model are employed as weights within the network, enabling the determination of the shortest path between any two stops. By incorporating predictive lateness information into the pathfinding process, our approach aims to enhance transportation efficiency and optimize travel routes in a way that will be transparent to PT users. 

The multiple key contributions of this study are the following:
\begin{itemize}
    \item We design a novel method that uses deep learning to generate weights for public transportation graphs based on smart card data. 
    \item We establish a data-driven approach to improve public transport routing without changing the stop locations or their order. This approach seamlessly creates short-term improvements for PT routes.
    \item We demonstrated the method's effectiveness by evaluating it in Tel Aviv showing it can improve 9\% of routes.
    \item The fact that only a small percentage of lines did change – suggests the method is of added value because we do not compromise the integrity of the entire PT network.
\end{itemize}

The rest of the paper is organized as follows: In Section  \ref{sec:rw}, we review related work on smart card usage, route planning, and shortest path algorithms. Section \ref{sec:methods} describes the use case, experimental framework, and methods used to develop the deep learning model and its usage. In Section \ref{sec:results}, we present the results of our model. Section \ref{sec:dis} discusses the findings' implications and the study’s limitations and presents our conclusions and future research directions.

\section{Related Work}
\label{sec:rw}
\subsection{Travel Behavior Analytics}
Travel behavior dynamics are valuable in various domains, especially for transportation planners.
By understanding travel behavior, it is possible to plan better PT routes,  predict demand and improve the quality of PT services \citep{briand2017analyzing}.
The classical way to study travel behavior is by conducting surveys to model users' travel patterns  \citep{stopher2007household}.
While travel behavior surveys are valuable tools, they are also expensive and inefficient for modeling large populations \citep{stopher2007household}.
On the other hand, automated fare collection (AFC) based data sources such as smart cards can generate millions of records and include historical data \citep{maeda2019detecting}.
AFC-based payment was introduced two decades ago, and it is today a popular PT payment option worldwide. 

Besides being a convenient payment option, AFC can produce all passenger's geocoded and timestamped boarding data \citep{faroqi2018applications, pelletier2011smart}.
Moreover, it can be used to generate data regarding line transfers and alightings of various PT vehicles.
Such information is priceless for a variety of use cases, including studying travel patterns, crowdedness, and measuring PT performance \citep{agard2007mining, bryan2007understanding,wang2011bus, alguero2013using,zhao2017spatio,li2018smart,jang2010travel,yap2020crowding}.

Studying travel behavior has various challenges, such as Origin-destination (OD) matrix estimation \citep{wang2011bus}.
The OD matrix models travel patterns between locations over different timeframes \citep{gordon2013automated, munizaga2012estimation, chu2008enriching}.
Until smart cards, OD matrices were generated based on surveys via sampling \citep{chen2016promises}. 
However, surveys could not represent the travel patterns of the entire population.
Smart cards-based OD estimations produced better OD estimation enabling more advanced and precise analytics \citep{chu2008enriching,  wang2011bus, munizaga2012estimation}.



\subsection{Machine Learning}
Traditional PT analysis methods could not utilize the full potential of the rise in the amount of available data and computational power, which served as a catalyst for the creation of various new methods and algorithms \citep{welch2019big}.  
Examples include \citet{agard2007mining}, who identified daily travel patterns by clustering user groups; \citet{bhaskar2014passenger} clustered passengers by applying a density-based spatial clustering application with noise (DBSCAN) algorithm. 

All these changes and advances have changed PT research, where new and more complex methods are used to create more comprehensive studies than ever before  \citep{welch2019big}. 
\citet{palacio2018machine} demonstrated that machine learning algorithms are more accurate than linear models for forecasting PT demand; 
\citet{traut2019identifying} identified dangerous PT stops by fusing AFC data with crime records. 
\citet{tang2022domain} showed that it is possible to use transfer learning for PT by using a base model trained on a different city.

PT issues have also been addressed using Deep Learning or Deep Neural Net (DNN) models. Examples of such deep learning implementations include inference of passenger employment status \citep{zhang2019deep}, forecasting passenger destinations \citep{jung2017deep, toque2016forecasting}, inference of demographics \citep{zhang2019deep2}, improving passenger segmentation \citep{chen2018traveler}, and predicting multimodal passenger flows \citep{toque2017short}. Regression problems using deep learning in PT include traffic flow \citep{huang2014deep, smith2002comparison}, traffic forecasting \citep{yu2017deep} and travel time estimation \citep{tran2020deeptrans}. Further reviews can be found in \citep{nguyen2018deep, wang2019enhancing, kumar2021applications}.

\subsection{Route Planning and Decision Support Systems}
PT network design (PTND) is one of the main challenges in PT improvement. One main PTND goal is planning ideal routes for passengers and operators \citep{bozyiugit2017public}. Route planning is a complex issue affecting public and private interests and involves many stakeholders, including political, technical and telecommunications \citep{cascetta2015new}. Route planning is also vital in reducing traffic load, with PT planners helping passengers to find a satisfactory route \citep{berczi2017stochastic}. At its core, there is a fundamental trade-off between passengers’ (customer) satisfaction and operator profitability \citep{ibarra2015planning,currie2010city}.  Customer satisfaction can be influenced by many factors such as terminal comfort, regularity, service coverage and frequency level.  The measurement of PT performance is thus a crucial tool for transport operators, especially when passengers have many factors influencing their satisfaction and loyalty \citep{karim2018measuring}. Multiple reports suggest that the current process is inefficient and a dire need exists for a more systematic computer-based approach \citep{soares2019adaptive}. Ideally, routes will vary depending on user preference, and while the shortest (distance) path is usually preferred it is not always optimal \citep{bozyiugit2017public}.  \citet{dib2017advanced} claimed that nowadays shortest-time travel is not the only parameter of interest for passengers; thus, an efficient multiobjective analysis must be incorporated into route planning. Route cost functions can vary by travel time, distance (i.e., shortest path), scenic value, etc. \citep{delling2009engineering}. Therefore, to comply with passengers' demands, as well as implementations in Smart City infrastructures, route planning must provide the corresponding solutions and intelligent features of the system \citep{spichkova2015formal}. Reviews can be found in \citep{ibarra2015planning,guihaire2008transit,mcleod2017urban,dodson2011principles, karim2018measuring}.

In particular, due to the complex data required by service planning and problem-solving, including route planning, there is a growing interest in using automated DSS \citep{vitale2014decision}. Such systems can provide better data-driven choice options for operators. DSS are automatic and efficient systems that support complex decisions and planning \citep{ocalir2016decision,vitale2014decision} and are also common in PT \citep{kazak2019decision, vitale2014decision}. They not only help automate and improve planners' working processes and maintain operator profitability but also increase passengers’ satisfaction levels \citep{moslem2020integrated}. A DSS gives an advantage over human decision-makers as it can determine the impacts of different decision variables on performance. In complex environments, decision-makers faced with large amounts of information suffer many limitations, which can be reduced with the help of a DSS \citep{van1998improving}. Further reviews can be found in \citep{sharda1988decision, eom1998survey, eom2006survey}.

\section{Methods}
\label{sec:methods}
In this study, we aim to show the potential of DSS-based solutions for reducing the lateness of PT arrivals. Suggesting a slight route change to a decision-maker can help improve the arrival times without changing the stops themselves. This makes it easier to incorporate changes in PT routes since it does not change the stops and still improves the quality of service. To this end, we developed Short-Term Routing Decision Support System (STR-DSS), a novel method suggestion for short-term route improvements. The method is based on finding routes that minimize the lateness of PT vehicle arrivals to stops as the primary target function, assuming that the stop locations are fixed. 
In the scope of this paper, lateness\footnote{Since shortest path algorithms cannot handle negative weights, we ignored early arrival observations which were subsequently removed from the smart card dataset.} will be defined as the set $L=\{l_{1,1,1}  \ldots  l_{l,s,d} | l\in \mathbb{R}^+ \}$ where  l is a PT line, s is a stop, d is the timestamp. 
Given a specific line, stop and timestamp lateness is defined as $l_{l,s,d}=at_{l,s,d}- st_{l,s,d}$ which is the difference between $at$ the arrival time and the $st$ scheduled time. Other optimization targets, such as the shortest distance also can be used. We chose to focus on lateness since it is a proper measurement of PT punctuality in Israel, which is the source of our data. 

STR-DSS is based on seven steps: First, we curated three PT datasets containing information that can help us characterize PT travel. Then, using these datasets, we constructed a graph representing the PT road network. Afterward, we extracted a set of features using the graph and additional data. Next, we constructed a deep learning MLP (multi-layer perceptron, simple neural network) based model to predict lateness based on geospatial and temporal features. To train the model, we used smart card data to determine the arrival times at PT stops and GTFS (General Transit Feed Specification) for the expected scheduled arrivals. Then, we constructed a set of graphs  that represents the PT road network at different time frames. Afterward, we use the model to assign weights to each edge in the constructed graphs based on the forecasted  lateness. Lastly, we used the graphs and shortest path algorithms to find faster PT routes. Our algorithm was developed under the constraint that the order of stops is permanent, given that stop locations are mandatory and cannot be altered in the short run. Our methodology is generic and can be used to optimize any given numeric metric of interest. 

\begin{figure*}
    \centering
  \includegraphics[width=0.98\textwidth]{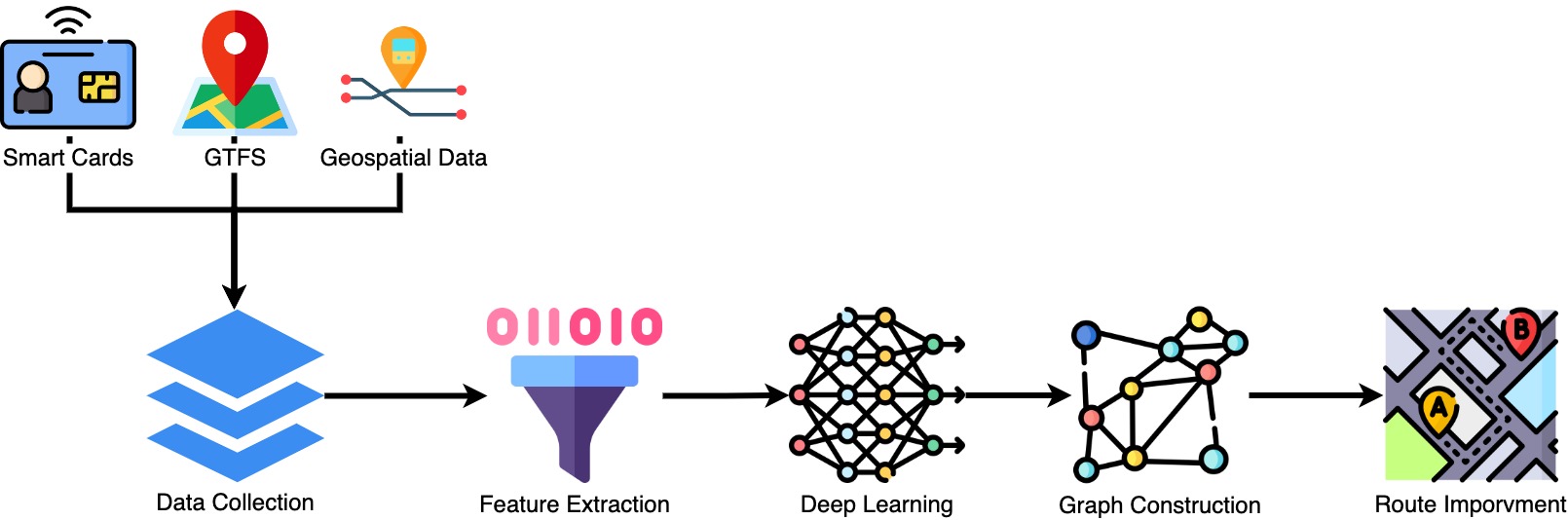}
  \caption[f]{Overview of the proposed methodology.\footnotemark}
  \label{fig:Methodology}
\end{figure*}

\subsection{Datasets and Data Preprocessing}
\label{subsect:data}
In this study, we utilized three datasets:
\begin{enumerate}
    \item 	\textbf{Smart card dataset} – "Rav Kav" is the Israeli national AFC system that applies the Transit Access Pass (TAP) protocol, allowing PT passengers to pay for their trip using their smart cards anywhere in the country. Rav-Kav operates a nationwide TAP IN for buses and rail that codes information on unique passenger identifiers, traveler types (such as student or senior travelers), boarding stops, boarding timestamps, fares, and discount attributes, and unique trip identifiers of the line at that time. For rail trips only, TAP OUT also records alighting stops and times. During 2018/9, for which our data is relevant, circa 2M boardings were recorded per day in the country.

    \item \textbf{GTFS dataset} - A GTFS feed, as described above, consists of rail/bus schedules, timetables, stops, and routes of every PT trip planned for every day of the month. In Israel since 2012, the GTFS feed has been regularly published daily online by the Ministry of Transport, providing schedules of 36 bus and rail operators, encompassing 7,800 route-direction-pattern alternatives served by 28,000 bus and rail stations. The GTFS feed aligns with the smart card data set as described below. The GTFS dataset was also used to enrich the feature space and to convert boarding stop records into an embedded numerical value.

    \item \textbf{Geospatial information} - a variety of geospatial attributes were derived from open municipal GIS databases (see Table \ref{tbl:features}).
\end{enumerate}

\begin{table}[]
\caption{The geographical features used by the deep learning model.} \label{tbl:features}
\begin{tabular}{|l|p{3.5cm}|}
\hline
\textbf{Feature name}            & \textbf{Explanation}                                            \\ \hline
Length                           & Length in meters, the total distance of segment                 \\ \hline
Number of   traffic lights       & Number of traffic lights in the segment                         \\ \hline
Number of PT   stops         & Number of public transport stops that are inside a   segment \\ \hline
Number of petrol   stations      & Number of petrol stations inside the segment                    \\ \hline
Number of public   parking spots & Number of public parking spots in the segment                   \\ \hline
Number of   private parking spots & Number of private parking spots in the segment                   \\ \hline
\end{tabular}
\end{table}

To obtain a dataset suitable for constructing the prediction model, we removed any record that lacked a boarding stop or a trip ID (a unique identifier of a trip provided by a specific and unique PT operator) from the smart card dataset. Next, we joined the smart card dataset with the GTFS dataset by matching the trip ID attributes. Lastly, we joined the geospatial dataset with the smart card dataset using the GTFS dataset, which contains each PT route's geographic coordinates.

\subsection{Feature Extraction}
To construct a model that can predict the lateness of a PT vehicle, we have to model the real-world environment in a way that can extract features that represent attributes that affect bus arrival. We look at PT movement as a graph where the PT moves along the edges of the graphs between nodes. Let $G=<V,E,\mathbb{A}	>$ where $V$ represents the set of points where PT may stop or change course (intersections and stops),  $E$ is the road segment between these points, and $\mathbb{A}=\{A_1, A_2 \ldots A_n\}$ is a set of n attributes that represents each road segment. That is, for each $e \in E$ , there is an attribute vector $\mathbb{A}(e) = (A_1(e),  \ldots , A_n(e))$ associated with $e$. 

To generate a graph $G_t$ that represents PT movement at the time of day t, we used the GTFS shape file to construct the graph for a specific city. This GTFS shape file lists each path of each scheduled PT trip.  A constraint was added that the PT must be able to pass through each road, i.e., the edge of the graph. Any route whose start or end vertex was part of the city network was also included (suburban lines). Additional constraints were added to verify each proposed route is feasible, such as the availability of turns. All of the above can be solved by generating a directed graph generated from the GTFS shape file. Since these are actual PT routes, they meet all these constraints.

Since the $\mathbb{A}(e)$ represents road segment $e$ it should have only attributes representing the road segment such as distance of the edge, number of traffic lights, etc., such features can be later used to predict information related to the edge. In this study, we have a set of six attributes that are later used as features in the deep learning model (see Table \ref{tbl:features}). Additionally, to the six geographical features extracted from graph $G$ we add an additional datetime feature. Although the geographical features are predictive, as asserted by \citet{shalit2022supervised}, they do not provide a good prediction for lateness, whereas the importance of temporal features (such as the planned schedules and boarding times) is higher. That is, the model's goal is not to predict the lateness of some routes but to understand the relation of features to lateness. For example, each traffic light adds 10 seconds to the lateness of a route, and 100 meters is also 10 seconds, so if the algorithm identifies a detour of less than 100 meters, it will prefer it to the traffic light. While this fact makes the model less precise, it is sufficient since it is never used directly, i.e., to predict the lateness of the smart card observations. We note that lateness is only an example target, other  tasks can also be used, including time to destination. As the graph has no explicit knowledge about the road segments, the model must extrapolate from the available data.

\begin{figure*}
\begin{subfigure}{.45\textwidth}
  \centering
  \includegraphics[width=0.9\columnwidth]{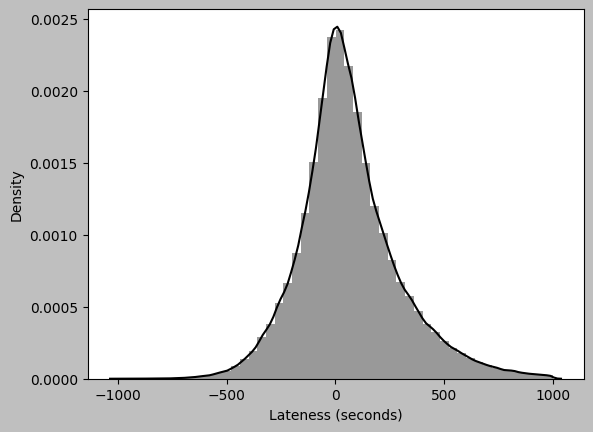}
  \caption{Beer Sheva}   \label{fig:LatenessBeer}
\end{subfigure}
\begin{subfigure}{.45\textwidth}
  \centering
  \includegraphics[width=0.9\columnwidth]{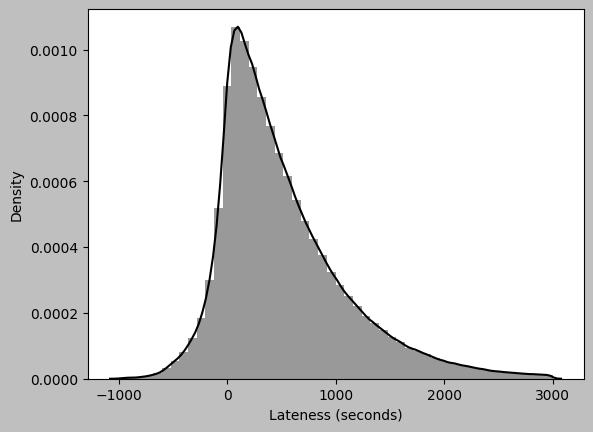}
  \caption{Tel Aviv}   \label{fig:LatenessTel}

\end{subfigure}
  \caption{Lateness in the cities of Beer Sheva and Tel Aviv in seconds.}
\label{fig:Lateness}
\end{figure*}
\footnotetext{This figure has been designed using images from Flaticon.com}

\subsection{Deep Learning Model Construction}

Lateness, the metric we are optimizing in this paper, was calculated using a combination of the smart card dataset and GTFS data (see Section~\ref{subsect:data}). We used the GTFS dataset to determine the scheduled arrival time of a PT vehicle to any stop (using the stop code feature) along each line and the smart card dataset to determine its actual arrival time. We then computed the difference between the actual and scheduled arrival times. Lateness happens when the difference is positive (see Figure~\ref{fig:Lateness}).

Our problem can be formalized as a regression problem for predicting the number of seconds (of the arrival time at a stop) using geospatial features and the time-of-day added relative to the expected scheduled arrival time. That, in turn, will predict the weight of each edge of the graph. We divided the features into two categories: First, numeric features, such as the length of the route, and second, categorical features, such as the period of the day. For real-time applications, it is possible to use numerical representations of time; however, for practical reasons, we generated a graph per time of day. Using part of the day creates five graphs that are easier to model and evaluate. It is possible to create more graphs, for instance, for each day of the week combined with the part of the day. Hence, we decided to handle categorical features separately. Embedding is a common method to handle categorical features with deep learning \citep{guo2016entity, arik2019tabnet,yang2014embedding, bordes2011learning}, also known as entity embedding. Since the model cannot train using textual categories (such as “morning” and “afternoon”), we converted our categorical features into vectors via embedding. 

\begin{figure}
    \centering
  \includegraphics[width=0.9\columnwidth]{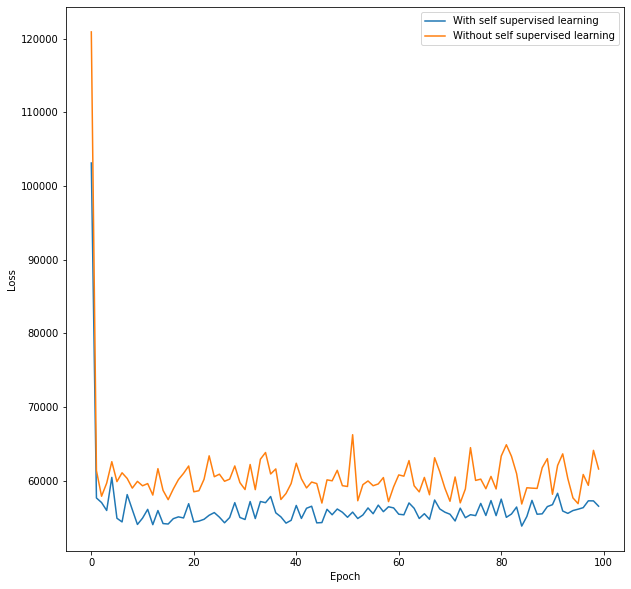}
  \caption{Loss of the deep learning model with and without supervised learning. }
  \label{fig:supervised-learning-loss}
\end{figure}

Additionally, embedding is also used to compare the categorical values. By comparing the final numbers assigned to each category by the model, we can see which category is more similar in the given task context. Moreover, since we have an extremely large number of observations, 2.5M, versus the limited number of route features we had available (see Table 1), to better learn from our data, we used the methodology of self-supervised learning \cite{hendrycks2019using}. This method is well-known for helping to stabilize models and increase their robustness by generating an Autoencoder (AE) where the decoder has the same architecture as the final neural network. Thus, in the AE training, the encoder layer learns to compress the data to represent it best (self-supervision) well. The output of the trained encoder is used as an input to train the MLP classifier, which should produce a more robust model. Additionally, such architecture allows us to pre-train the model and perform transfer learning between cities. As shown in Figure \ref{fig:supervised-learning-loss}, the self-supervision model converged faster and with smaller loss, i.e., the deep learning network is more accurate \citep{choromanska2015loss}. We trained the deep learning model using RMSE as a loss function, meaning the root mean square error between predictions and actual values, i.e., lateness.

\subsection{Route Suggestion}
To suggest faster routes for PT, we look at the problem as a graph search problem. Formally, let $G_t^`   =<V,E,\mathbb{A},w>$  be the graph where $G_t^`$ is a graph representing the PT road network at time frame t which we treat as the part of the day the embeddings represents. The graph is constructed precisely as graph G only with weight $w$, which is the predicted lateness value by the deep learning model. Once the graph is generated, it is static, and each set of k stops will generate a single deterministic route. In this task, we used the well-known Dijkstra shortest path algorithm \cite{bozyiugit2017public}. Dijkstra is used to suggest the fasted route based on the edges' weights.

\subsection{Case Study}

We considered two case studies: First, we piloted the methodology on a smaller network of Beer Sheva, the largest city in the Southern part of Israel with about 200,000 inhabitants. This network includes 30 routes consisting of 100,014 vertices and 129,862 edges. However, since lateness is not too severe (see Figure~\ref{fig:LatenessBeer}) and the number of bus routes is small, few route modifications were identified. Next, we tested the methodology on the much larger network of Tel Aviv, where lateness is much more acute (see Figure~\ref{fig:LatenessTel}). This network includes 543 routes (including urban and suburban lines) consisting of 195,222 vertices and 224,842 edges. This result is not surprising as the city is the country's leading commercial and financial hub and the core of a metropolitan region of over 3M inhabitants, which is much more prone to traffic congestion.

\begin{figure}
    \centering
  \includegraphics[width=0.9\columnwidth]{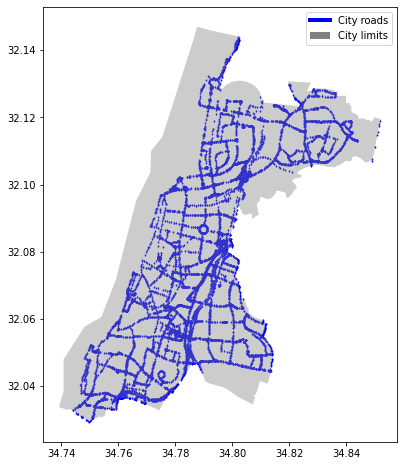}
  \caption{PT graph of the city of Tel Aviv.}
  \label{fig:TA_graph}
\end{figure}

\section{Results}
\label{sec:results}
It is important to note that the developed tool was designed as a decision support system to assist but not replace human planners. The aim is to rapidly generate a suggested PT route as a benchmark and create ideas for further discussion. 

First, we analyzed the PT network of the city of Beer Sheva. However, in most cases, the model did not suggest changing the  original routes. Accordingly, to demonstrate the power of our algorithm, we analyzed the PT graph of Tel Aviv, which has a much richer road network. Due to these reasons, in this section, we focused on the network for the city of Tel Aviv (see Figure \ref{fig:TA_graph}).

\begin{figure}
    \centering
  \includegraphics[width=0.98\columnwidth]{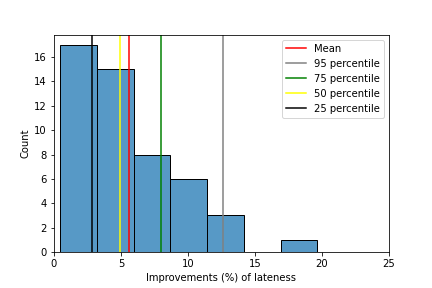}
  \caption{Distribution of routes and percentiles by improvements in the percentage of lateness}
  \label{fig:dist}
\end{figure}

Our model improved 50 routes in Tel Aviv, which is 9.2\% of the routes. Figure \ref{fig:dist}, presents the distribution of routes by the percentage of improvement in lateness, which is skewed to the right. Evidently, most of the improvements are minor, with over 50\% less than 5\% change. In the eyes of the transit planner, these can probably be disregarded. However, some of the routes show much more significant changes with the 95\% percentile above 12\% improvement. It is these candidates that the transit planner should now scrutinize professionally. Note that the cutoff when to intervene is in the hands of the planner to decide. The DSS only provides the list of candidates and flags those with higher chances to be improved but does not aim to replace the professional's eye.

\begin{figure*}
\begin{subfigure}{.49\textwidth}
  \centering
  \includegraphics[width=0.9\columnwidth]{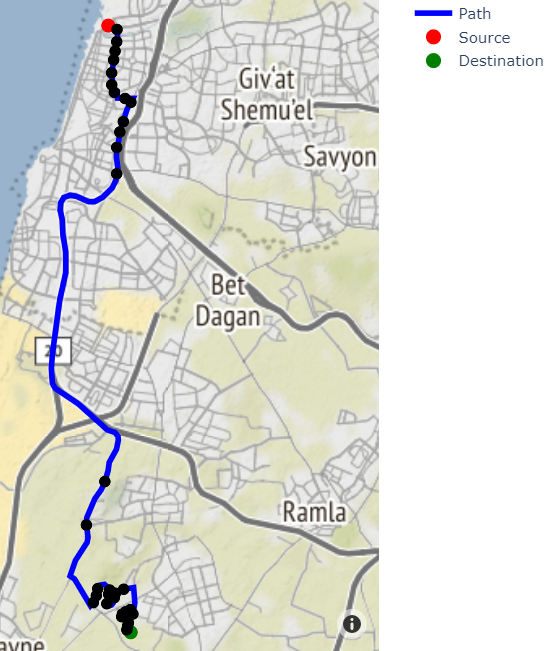}
  \caption{Original route}
\end{subfigure}
\begin{subfigure}{.49\textwidth}
  \centering
  \includegraphics[width=0.9\columnwidth]{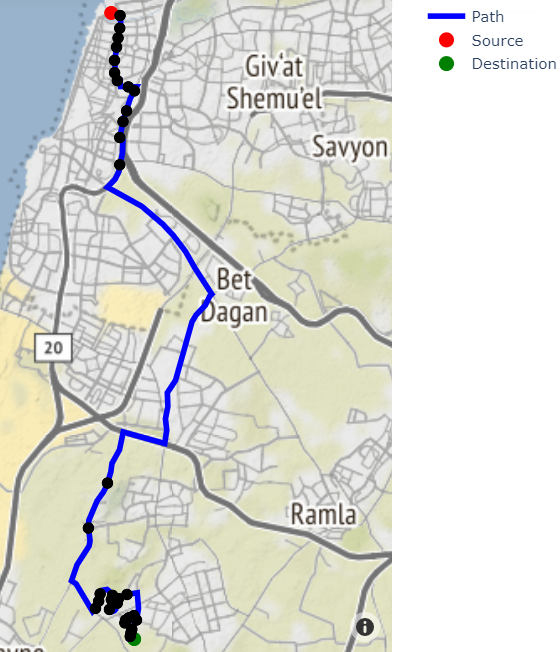}
  \caption{Proposed route}
\end{subfigure}
  \caption{Example \#1: The proposed route is shorter than the original route.}
\label{fig:ayalon-route}
\end{figure*}

We now show two examples that in our mind showcase how the model can improve the PT's performance. Our first example is Route 16272 (see Figure \ref{fig:ayalon-route}), with 2\% improvement in expected lateness, which connects the southern city of Rehovot and central Tel Aviv. We can see that the proposed route is quite different. Especially interesting is that the original route used "Road 20"–the "Ayalon" Highway (South)–which suffers from traffic congestion throughout the day. Naturally, the model could never know that explicitly, which asserts that the suggested change is unlikely trivial. To understand whether the suggested route is also a better one, we need to examine its attributes (see Table \ref{tbl:exmpl}). The latter clearly shows that the suggested route is shorter and superior according to the different features the model uses as inputs. It is also interesting that the improved route follows a dedicated bus lane that did not exist on the Ayalon Highway. In addition, the improved route traverses another major city Rishon Le-Zion which helps to make the route more attractive to the operator by attracting more passengers.  

\begin{table}[]
\caption{Comparison of routes attributes of Example \#1} \label{tbl:exmpl}
\begin{tabular}{|l|l|l|}
\hline
\textbf{Route 11277}            & \textbf{Original} & \textbf{Proposed} \\ \hline
Length (meters)                 & 30789             & 28370             \\ \hline
Number of traffic lights        & 33                & 33                \\ \hline
Number of PT terminals          & 0                 & 0                 \\ \hline
Number of petrol stations       & 1                 & 1                 \\ \hline
Number of public parking spots  & 14                & 14                \\ \hline
Number of private parking spots & 251               & 244               \\ \hline
\end{tabular}
\end{table}

\begin{figure*}
\begin{subfigure}{.49\textwidth}
  \centering
  \includegraphics[width=0.98\columnwidth]{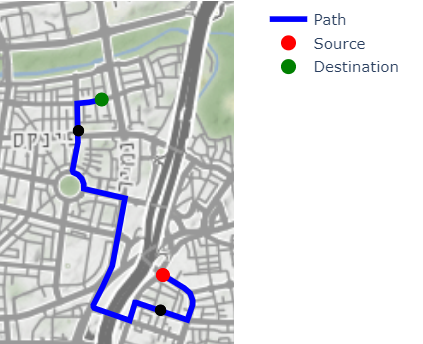}
  \caption{Original route.}
\end{subfigure}
\begin{subfigure}{.49\textwidth}
  \centering
  \includegraphics[width=0.98\columnwidth]{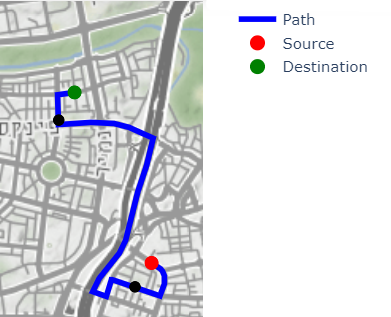}
  \caption{Proposed route}
\end{subfigure}
  \caption{Example \#2: The proposed route is longer than the original route.}
\label{fig:namir-route}
\end{figure*}

A key question is whether the model accidentally chose the presented shortest paths. If the answer is positive, then this would be of no real interest. Therefore, we show a second example (see Figure \ref{fig:namir-route}), with 10\% improvements in expected lateness, where the suggested route is longer, but other attributes have improved, eventually leading to a better route (see Table \ref{tbl:exmpl2}). The original route passed on "Namir Road," a main Northbound thoroughfare, whereas the proposed route passes through the "Ayalon" Highway (North). It is easy to see that the proposed route passes through fewer signalized intersections and has fewer turns, resulting in shorter travel times.

\begin{table}[]
\caption{Comparison of routes attributes on second example} \label{tbl:exmpl2}
\begin{tabular}{|l|l|l|}
\hline
\textbf{Route 11900}            & \textbf{Original} & \textbf{Proposed} \\ \hline
Length (meters)                 & 4418              & 4666              \\ \hline
Number of traffic lights        & 12                & 8                 \\ \hline
Number of PT stops          & 0                 & 0                 \\ \hline
Number of petrol stations       & 0                 & 0                 \\ \hline
Number of public parking spots  & 3                 & 1                 \\ \hline
Number of private parking spots & 29                & 27                \\ \hline
\end{tabular}
\end{table}

\begin{figure}
    \centering
  \includegraphics[width=0.98\columnwidth]{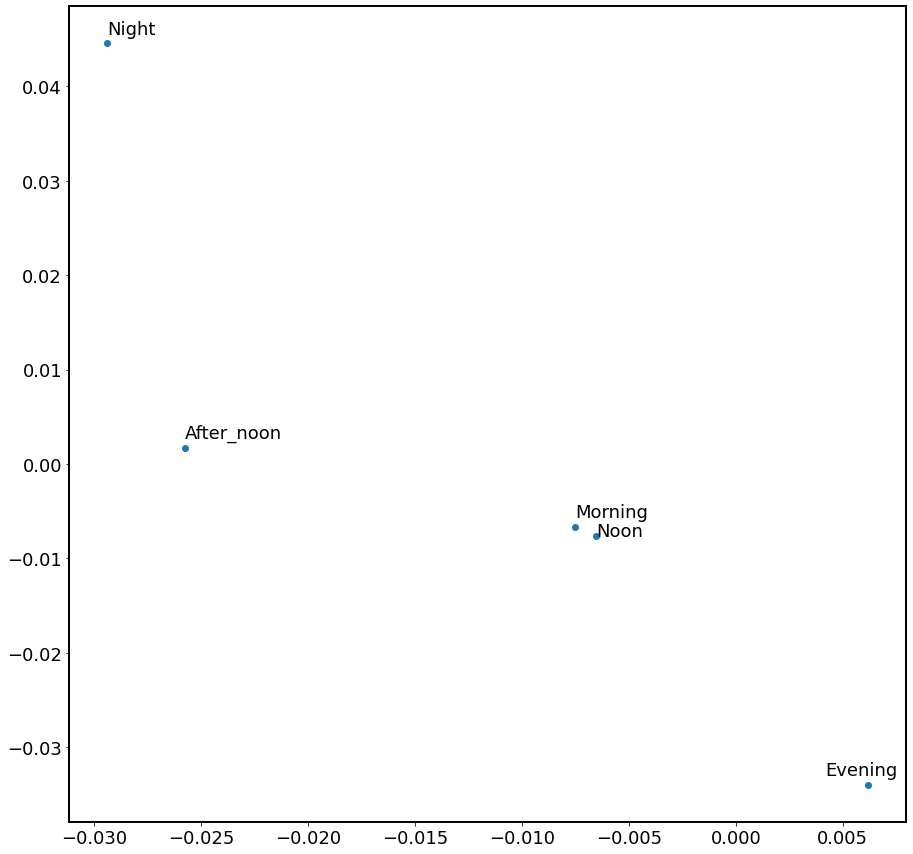}
  \caption{Learnt embedding of time-of-day in Tel Aviv data.}
  \label{fig:embedding}
\end{figure}

The implications of the learned embedding can be seen in Figure \ref{fig:embedding}. The graph presents PCA (Principle Components) transformation of the embedding space, so while the actual number (x/y axis) has no meaning, closer points are more similar. It is interesting to see that the Morning and Noon periods are extremely similar. In contrast, Evening and Night are highly dissimilar. This result means that for the model, the task of predicting lateness in the morning and at noon times is similar, probably due to the effect of traffic patterns that are more similar for a workaround and work-related traffic. In contrast, afternoon, evening, and night are probably dissimilar as the traffic patterns are quite different. The afternoon is more congested for homebound traffic, while evening traffic is more associated with shopping and entertainment. Nighttime is almost always free-flowing and uncongested. Overall, the results seem commonsensical. 

\section*{Discussion}
\label{sec:dis}
Improving PT routes has a high value in our society with growing urbanization and increased demand for precise transportation. In this work, we presented a novel model for a possible short-run PT route planning DSS. The model is based on a deep learning model for minimizing expected delays and harvesting smart card, geospatial, and GTFS data. The DSS is based on producing an alternative graph of proposed routes. The model was trained on the PT network of the city of Tel Aviv. The following points are important to note:
First, the model is a supporting tool for assisting planners but is never intended to replace them. That is, this is not artificial intelligence (AI); rather, it is expected planners or operators carefully scrutinize the proposals to see if they fit their needs. Second, we proposed a novel data-driven automated method to improve routes based on minimizing expected delay. This method is capable of optimizing various metrics that can be calculated from smart card data or geospatial data, such as lateness (as was shown here), probability of lateness (in percentages), shortest distance, minimum exposure (e.g., in the case of COVID-19), etc. These attributes allow for fast computation and harvesting of relevant data, which play a critical role in designing better PT routes.

Second, we created five graphs, each representing another part of the day. We have created only five graphs to make them more straightforward for training and evaluation. However, it is logical to assume that creating more fine partitioning to graphs should produce better routes. For instance, traffic patterns are entirely different on weekends than on weekdays when people leave the house based on school and work hours. We believe that creating graphs for smaller time frames would create more accurate predictions and a more complex model. Decision makers would have to define at which points in time it is more important to them to get better performance at the cost of complexity.

Third, the routes' quality highly depends on the graph's quality, i.e., cannot make improvements and changes if the graph is not sufficiently detailed. That is, if a route is not elaborate enough and different from the quickest path, it is unlikely that the model will suggest relevant detours and changes. It is important to note, again, even with the last two limitations, that this tool does not need to change routes but can help to create new routes. Since most of the routes we tested returned with the same layout, meaning instead of spending scarce resources on significant route designing, the planner can quickly obtain an answer on the needs for modifications, saving both time and costs. In the case of all routes in Beersheva and most in Tel Aviv, the model did not change the original route. This outcome highlights the advantages of the suggested approach as a DSS where, in most cases, it will quickly provide a result (only a few seconds once the initial graph is ready). Since the model predicted the same route as the original one in most cases, we believe it asserts its predictive power and potential as a decision support system. 

Fourth, the potential of the DSS tool is highly correlated with the generated graph and the quality of geospatial features. Better features will yield a more predictive model, which will yield more optimized routes. Temporal factors can be better granulated, e.g., a level of weekdays can be easily added, i.e., a route for Monday morning could well defer from that of Friday morning. Additionally, explicit domain knowledge can be incorporated in the graph, like the availability of PT routes, the number of lanes on the road, traffic volume (vehicles per hour), level-of-service (volume to capacity ratio) per time of day, etc. This knowledge body, in turn, will result in a more accurate graph and suggested routes.
Last, the suggested methodology can have other important implementations. For instance, to the best of our knowledge, the very well-known Traveling Salesman Problem (TSP) was researched mainly on shortest-distance solutions \citep{gutin2006traveling}. However, with our methodology, the graph weights can be modified, and improved solutions can be applied. More specifically, while we used a fixed order of stops to optimize the routes, the TSP solver can also determine the optimal order of the k stops.
Several limitations of the research are noteworthy. First, the model is suboptimal since we can only use generic features such as spatial features. The model can be improved if more features can be derived that we did not apply, such as the width of the road, number of lanes, number of vehicles per time of the day, etc. Additionally, the graph we generated was very generic, based on GTFS; this was done to show that the method is very robust and its concept should work in various conditions. Nevertheless, a more informative graph should likely yield better results. Furthermore, we did not use the information on the day of the week/ weekend, which could well generate different recommendations. Also, we believe this method can be used as an analysis tool to measure route efficiency or even as a stop-removal method. For instance, by removing one stop and using the model to search for a faster route, we can measure the time of a specific stop.

\section*{Conclusion}
In this work, we present a novel DSS method for short-term improvements of PT routes.
To this end, we designed a method that combines deep learning and graph theory-based techniques. The model was designed to generate weights for every road segment in a way that it could be converted into a graph search problem. This enables us to search for a faster path between two subsequent stations without changing stations.
Additionally, we can use the method to search for faster ordering of stations.
We evaluated the method's effectiveness on Tel Aviv PT lines and found improved options for 9\% of the routes.
The fact that the algorithm only suggests changing the routes of a relatively small percentage of the lines strengthens its value as DSS since it does not compromise the integrity of the entire PT network.
We hope this method will inspire PT operators to use DSS to improve the quality of service to the public constantly.

\phantomsection
\section*{Acknowledgments}This research was supported by the Ministry of Science \& Technology, Israel, and The Ministry of Science \& Technology of the 'People's Republic of China (Grant No. 3-15741). Special thanks to Data Scientist Raz Vais, advisor to the Israeli National Public Transport Authority, for help in obtaining and processing the smart card data.

\bibliographystyle{unsrtnat}
\bibliography{sample-base}










\end{document}